\documentclass[runningheads]{llncs}
\usepackage{graphicx}

\usepackage{tikz}
\usepackage{comment}
\usepackage{amsmath,amssymb}

\usepackage{multicol}
\usepackage{color}
\usepackage{booktabs}

\usepackage[accsupp]{axessibility} 

\begin{document}
\pagestyle{headings}
\mainmatter
\def\ECCVSubNumber{22} 

\title{Racial Bias in the Beautyverse: 
Evaluation of augmented-reality beauty filters}

\titlerunning{ECCV-22 submission ID \ECCVSubNumber} 
\authorrunning{ECCV-22 submission ID \ECCVSubNumber} 
\author{Anonymous ECCV submission}
\institute{Paper ID \ECCVSubNumber}

\titlerunning{Racial Bias in the Beautyverse}

\author{Piera Riccio\inst{1,2}\orcidID{0000-0001-8602-8271} \and
Nuria Oliver\inst{1,2}\orcidID{0000-0001-5985-691X}}

\authorrunning{P. Riccio and N. Oliver}

\institute{ELLIS Alicante Foundation\and
\email{\{piera,nuria\}@ellisalicante.org}}

\maketitle

\begin{abstract}
This short paper proposes a preliminary and yet insightful investigation of racial biases in beauty filters techniques currently used on social media. The obtained results are a call to action for researchers in Computer Vision: such biases risk being replicated and exaggerated in the Metaverse and, as a consequence, they deserve more attention from the community.
\end{abstract}

\section{Introduction}

The Metaverse may be conceived as the culmination of the digitization of our lives and our society, leveraging key technological advances in fields, such as Augmented, Virtual and Mixed Reality and Artificial Intelligence. From a societal perspective, the Metaverse is considered to be the next stage in the development of current social media platforms \cite{metaverse2040}. In this regard and similarly to what happens on social media, the broad adoption and use of the Metaverse by potentially billions of users poses significant ethical and societal challenges, including the need to develop an inclusive environment, respecting the diversity of its users \cite{lee21}. Thus, it is of paramount importance to ensure that the enabling technologies of the Metaverse do not create, replicate or even exacerbate patterns of discrimination and disadvantage towards specific groups of users: fairness and diversity should be at the foundation of its development \cite{woodruff18}. 

In this paper, we focus on diversity in self-representation in the Metaverse. Specifically, we study the existence of implicit racial biases behind the Augmented-reality (AR)-based selfie beautification algorithms that are pervasive in social media platforms. We leverage Computer Vision techniques to perform such a study and argue that current user behaviors observed in today's social media platforms may be analyzed as an anticipation of what will happen in the Metaverse. We refer to the set of these new self-representation aesthetic norms as the \textit{Beautyverse}. Note that existing and under-studied biases in the \textit{Beautyverse} could lead to harmful appearances in avatar representations  in the Metaverse \cite{divine,neely17}. Thus, we highlight the importance of coupling the proposal of novel technical contributions for the Metaverse with a comprehensive, multidisciplinary study of their societal implications.

\section{Related Work}

Self-representation in the digital space is a key factor in online social media platforms that will also shape the social interactions in the Metaverse \cite{552}. In current social media platforms, self-representation is expressed through selfies (photos of the self); in the Metaverse, selfies are translated to avatars, which are common in other types of online environments, such as video games. Behind the self-representation through avatars there is a will to create an ideal version of the self \cite{mummendey95,higgins87,manago08}, including both the personality \cite{bessiere07} and the appearance \cite{ducheneaut09,messinger08} of the avatars. 

Our research focuses on the improvements on the appearance. In recent years, AR-based selfie filters that \emph{beautify} the original faces have become very popular on social media platforms \cite{techreview}. Previous work has linked these filters to the definition and adoption of new facial aesthetics \cite{2021filtering}, with significant social and cultural impact, such as an exponential increase in teen plastic surgeries \cite{khunger2021} and mental health issues \cite{abijaoude2020}. These filters have been widely criticized for perpetuating racism \cite{huffpost}, since the beautifying  modifications applied to the original faces include lightening the skin tone, reducing the size of the nose and making the eyes bigger and lighter, which imply that people should look \textit{whiter} to be considered beautiful \cite{complex}. In addition, the Eurocentrism of beauty filters is also shown in the \textit{colonization of ethnic features as an aesthetic} \cite{nylon}, accepting certain features only when applied on the faces of white people, and rejecting them in other cases \cite{techreview2}. Moreover, the perpetuation of discriminating and racist beauty ideals in the \textit{Beautyverse} could lead to significant cultural damage and dysmorphia when applied to avatars, given that avatars do not require an underlying physical reality, as selfies do. 

In this paper, we aim to shed light on the existing dynamics that create the aesthetic norms and ideals of the \emph{Beautyverse}. We address this challenge from a computational perspective, with the intent of bringing sociological and anthropological research questions to the Computer Vision community.

\section{The Implicit Racial Bias in AR-based Beauty Filters}

\begin{figure}[ht]
    \centering
    \includegraphics[width=\textwidth]{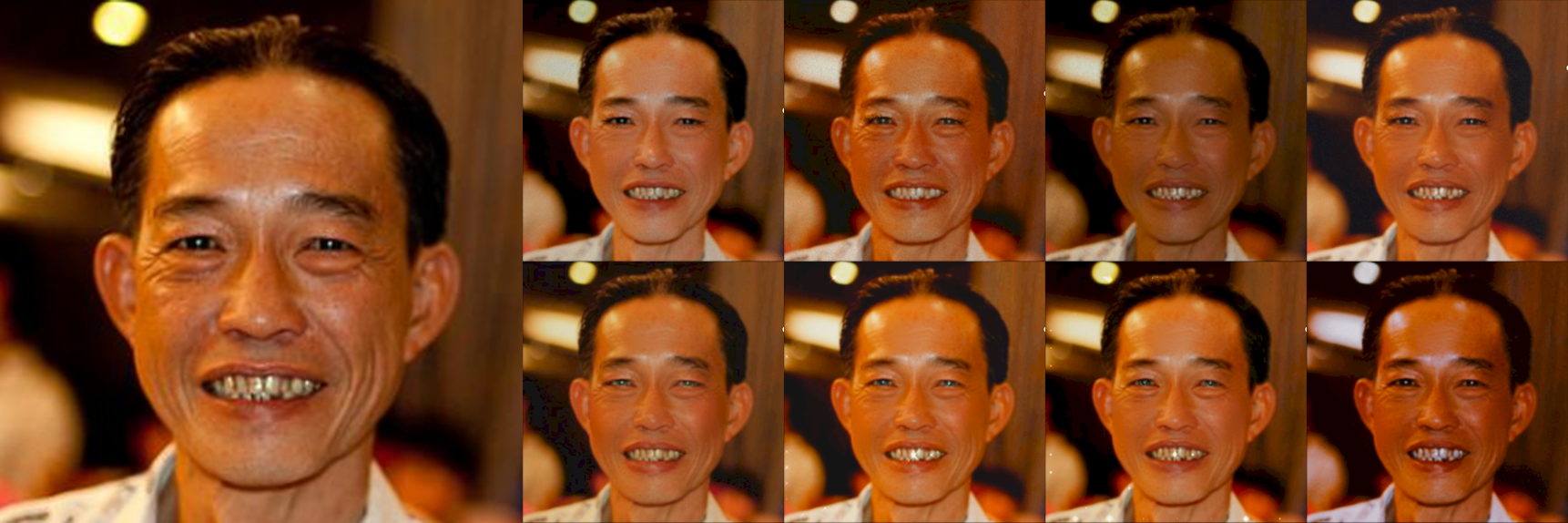}
    \caption{Example of the eight different beauty filters applied to the left-most image from the \textsc{FairFace} dataset \cite{fairface}.}
    \label{fig:filters}
\end{figure}

The aim of our research is to leverage Computer Vision techniques to understand key characteristics of the \textit{Beautyverse}, with a special focus on racial biases. We report results of preliminary analyses on the \textsc{FairBeauty} dataset \cite{riccio2022}. This dataset has been designed to enable the study of the implications of AR-based beauty filters on social media. It was built by beautifying the faces of the \textsc{FairFace} dataset \cite{fairface} via the application of eight AR-based filters available on Instagram. The filters were chosen based on their popularity and were directly applied on the images of the \textsc{FairFace} dataset \cite{fairface}, as illustrated in Figure \ref{fig:filters}. 

Given the diversity in the \textsc{FairFace} dataset, \textsc{FairBeauty} enables the study of the differential impact of beautification filters on faces of different ages, genders and races. Previous work has shown that the applied beautification filters homogenize the appearance of the faces, and thus increase the similarity among individuals \cite{riccio2022}. It has also been shown that the level of homogenization does not impact the performance of state-of-the-art face recognition models. In summary, these filters modify the faces so they conform to the \emph{same} beauty standard while preserving the identity of the individuals \cite{riccio2022}.

In this paper, we study the existence of implicit racial biases in the beautification filters. Specifically, we answer the following research question: do AR-based beautification filters encode a canon of beauty of white people? We describe next the experimental setup to address our research question.

\subsection{Experimental Setup}
In our experiments, we investigate whether beauty filters implicitly make beautified individuals of all races look \textit{whiter}. To tackle this question, we leverage two state-of-the-art race classification algorithms: DeepFace \cite{deepface} and FairFace \cite{fairface}. The faces in \textsc{FairFace} are labeled according to seven different racial groups, namely: Black, East Asian, Indian, Latino Hispanic, Middle Eastern, Southeast Asian, and White. In our experiments, we randomly sample a subset of 5,000 faces for each race. We compare the performance of the race prediction algorithms on the face images from \textsc{FairFace} \cite{fairface} and the corresponding beautified version in \textsc{FairBeauty} \cite{riccio2022}.

The first race classification model used in the experiments is DeepFace \cite{deepface}, which is a lightweight face recognition and facial attribute analysis framework. It is available in the \texttt{deepface} Python library and it is based on different face recognition models. The framework is trained to recognize four attributes: age, gender, emotion and race. In this paper, we focus on race recognition with pre-training on the VGGFace2 dataset \cite{vgg-face}. The second race classification model is the one released with the publication of the \textsc{FairFace} dataset \cite{fairface}. In this case, the race predictor is based on ResNet34.

\subsection{Results}

In this section, we present the race classification results on original and beautified faces. Tables \ref{tab:deepface} and \ref{tab:ff} depict the average predicted value of the label \emph{White} (mean and standard deviation) in each racial group by the DeepFace and FairFace algorithms, respectively. The values are averaged over the 5,000 randomly selected faces for each of the 7 races. The right-most column on the tables presents the prediction loss, i.e. the difference in the race classification performance between the original and the beautified datasets.

As shown on the Tables, the predicted value of the label \emph{White} significantly increases in the beautified faces of \emph{all races} when compared to the original, non-beautified images. Moreover, there is a significant loss in the performance of the race classification algorithm when applied to the beautified faces of most races except for the images labeled as \emph{Whites}, whose performance increases in the beautified version of the original faces. In other words, there is a larger probability to classify the beautified faces --independently of their race-- as white. 

\begin{table}
\caption{Race classification results of the DeepFace algorithm, applied to 5,000 images. The first and second columns depict the predicted value of the label \textit{white} in the original \textsc{FairFace} and the beautified \textsc{FairBeauty} datasets, respectively; the third column contains to the race prediction loss, i.e. the difference in the race classification performance between the original and the beautified datasets.}
\label{tab:deepface}
\centering
\vspace{6pt}
\def\arraystretch{1.2}
\begin{tabular*}{0.90\textwidth}{lccc}
\toprule
& \multicolumn{1}{c}{Original} \hspace{11pt} & \multicolumn{1}{c}{Beautified} \hspace{11pt} & \multicolumn{1}{c}{True Prediction Loss}\\
\midrule
Race: &  &  & \\
\midrule
\hspace{22pt} Black  & $4.35 \pm 0.18$ & $7.23 \pm 0.23$ & -7.36\%\\
\hspace{22pt} East Asian & $10.84 \pm 0.27$ & $14.00 \pm 0.29$ & -7.83\%\\
\hspace{22pt} Indian & $10.07 \pm 0.21$ & $15.09 \pm 0.26$ & -7.73\%\\
\hspace{22pt} Latino & $20.71 \pm 0.28$ & $26.85 \pm 0.33$ & -4.00\%\\
\hspace{22pt} Middle Eastern & $27.73 \pm 0.31$ & $35.35 \pm 0.35$ & -4.05\% \\
\hspace{22pt} Southeast Asian \hspace{11pt} & $8.88 \pm 0.23$ & $12.48 \pm 0.26$ & -11.23\%\\
\hspace{22pt} White & $53.46 \pm 0.44$ & $57.89 \pm 0.44$ & +4.43\% \\
\bottomrule
\end{tabular*}
\end{table}

\begin{figure}
    \centering
    \includegraphics[width=0.5\columnwidth]{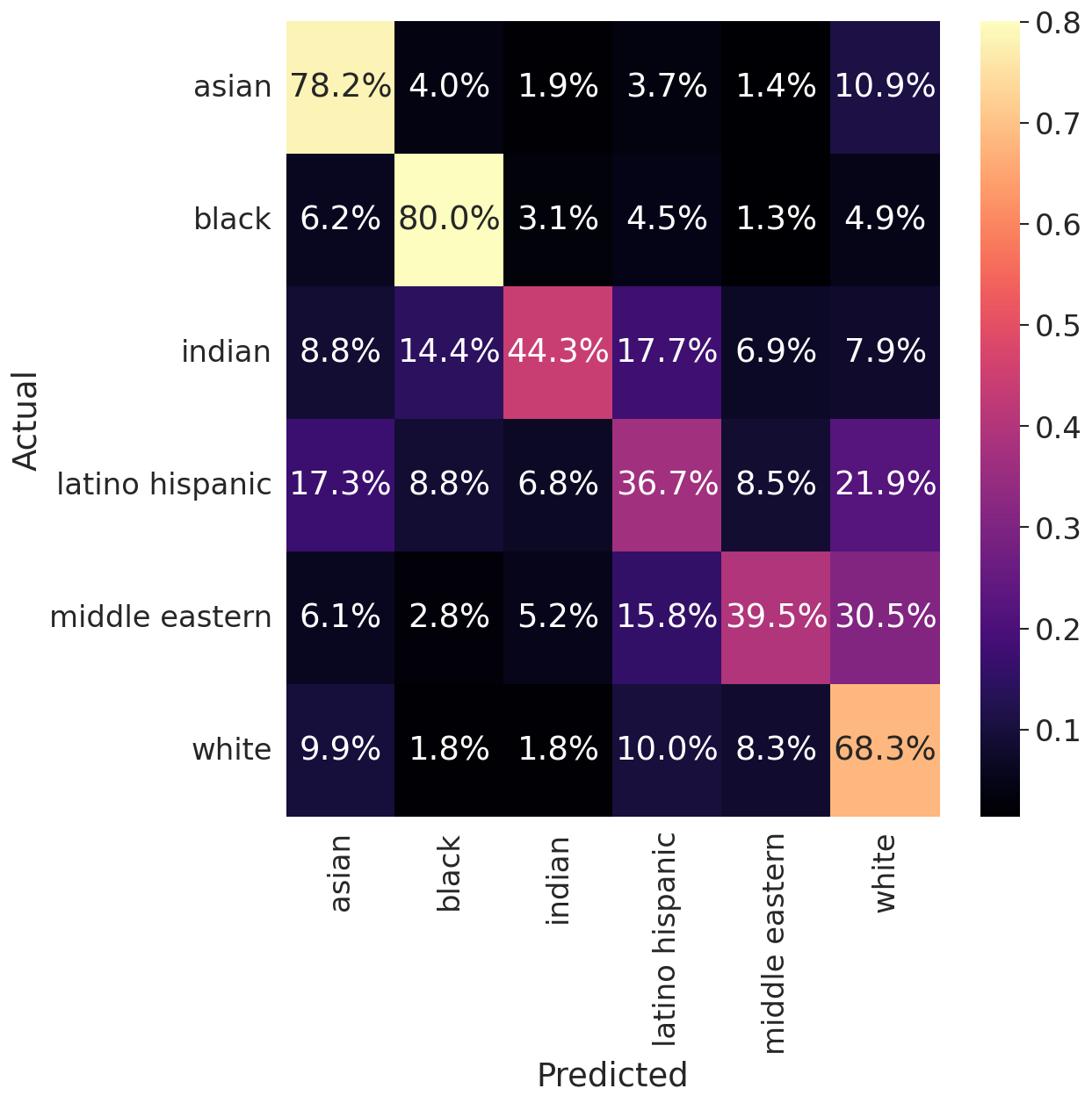}
    \includegraphics[width=0.5\columnwidth]{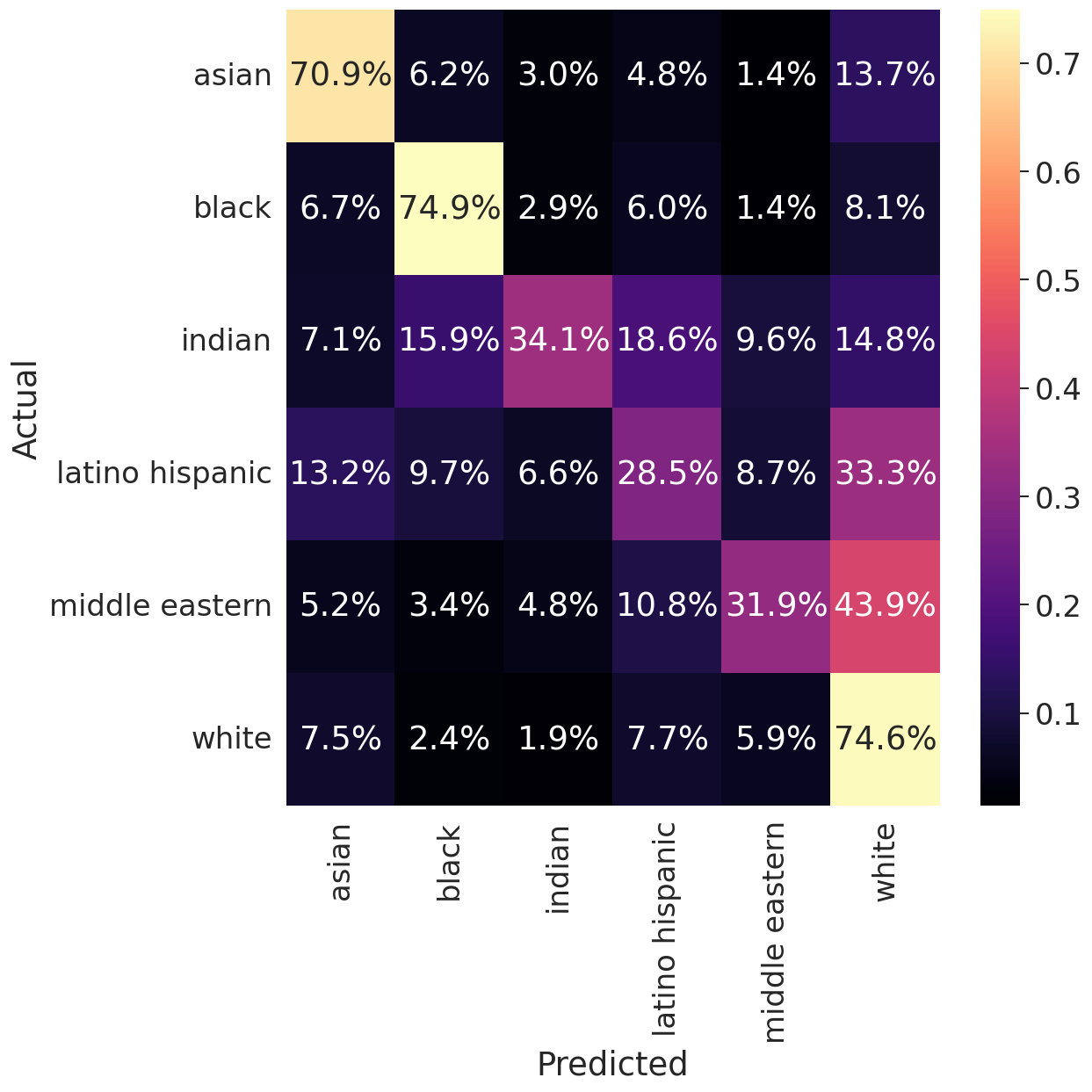}
  \caption{Confusion matrices of the race prediction, obtained through DeepFace \cite{deepface}. Left-hand side refers to the original images, right-hand side refers to the images after beautification.}
  \label{fig:deepface}
\end{figure}

\begin{figure}
    \centering
    \includegraphics[width=0.49\columnwidth]{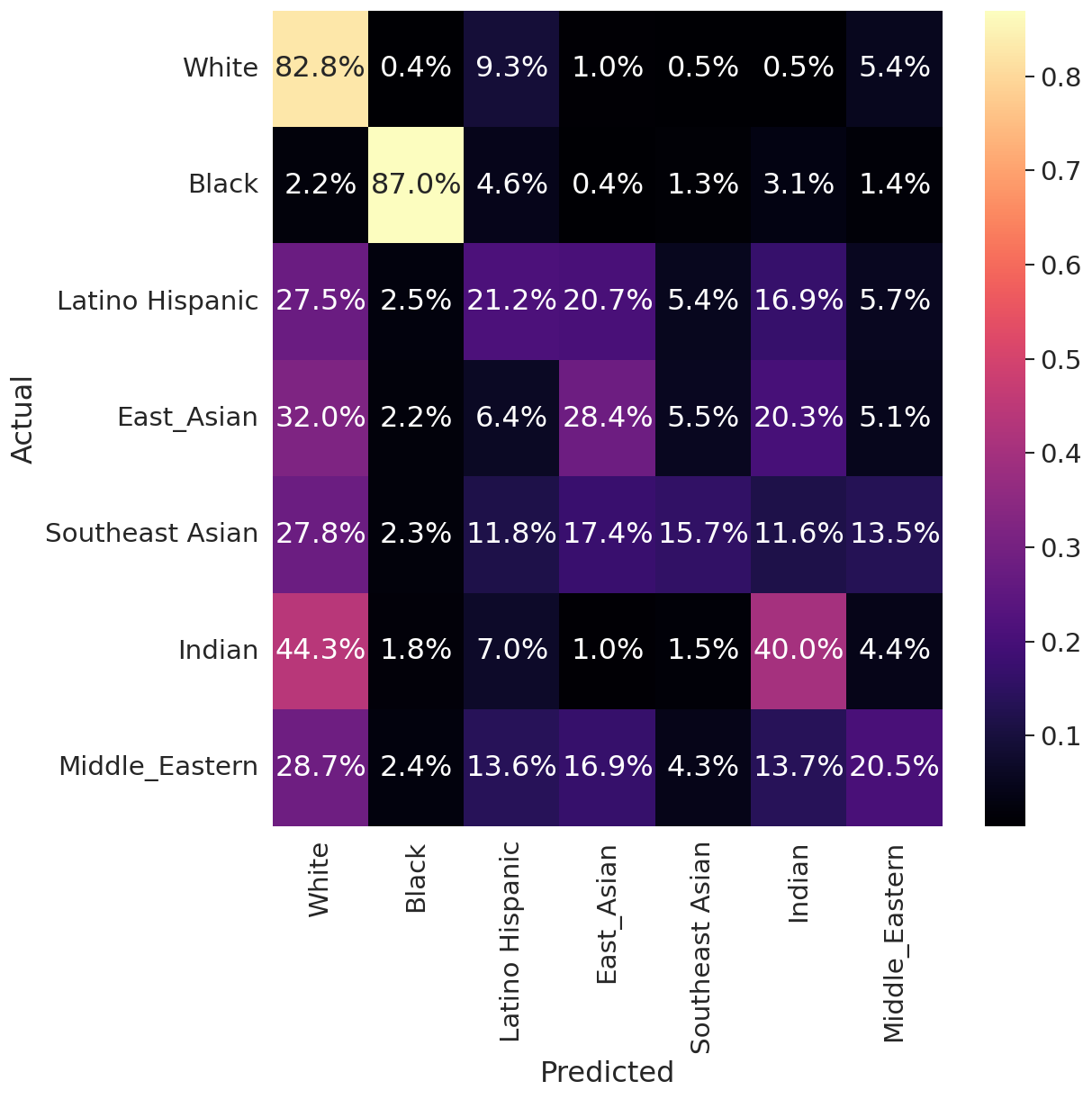}
    \includegraphics[width=0.49\columnwidth]{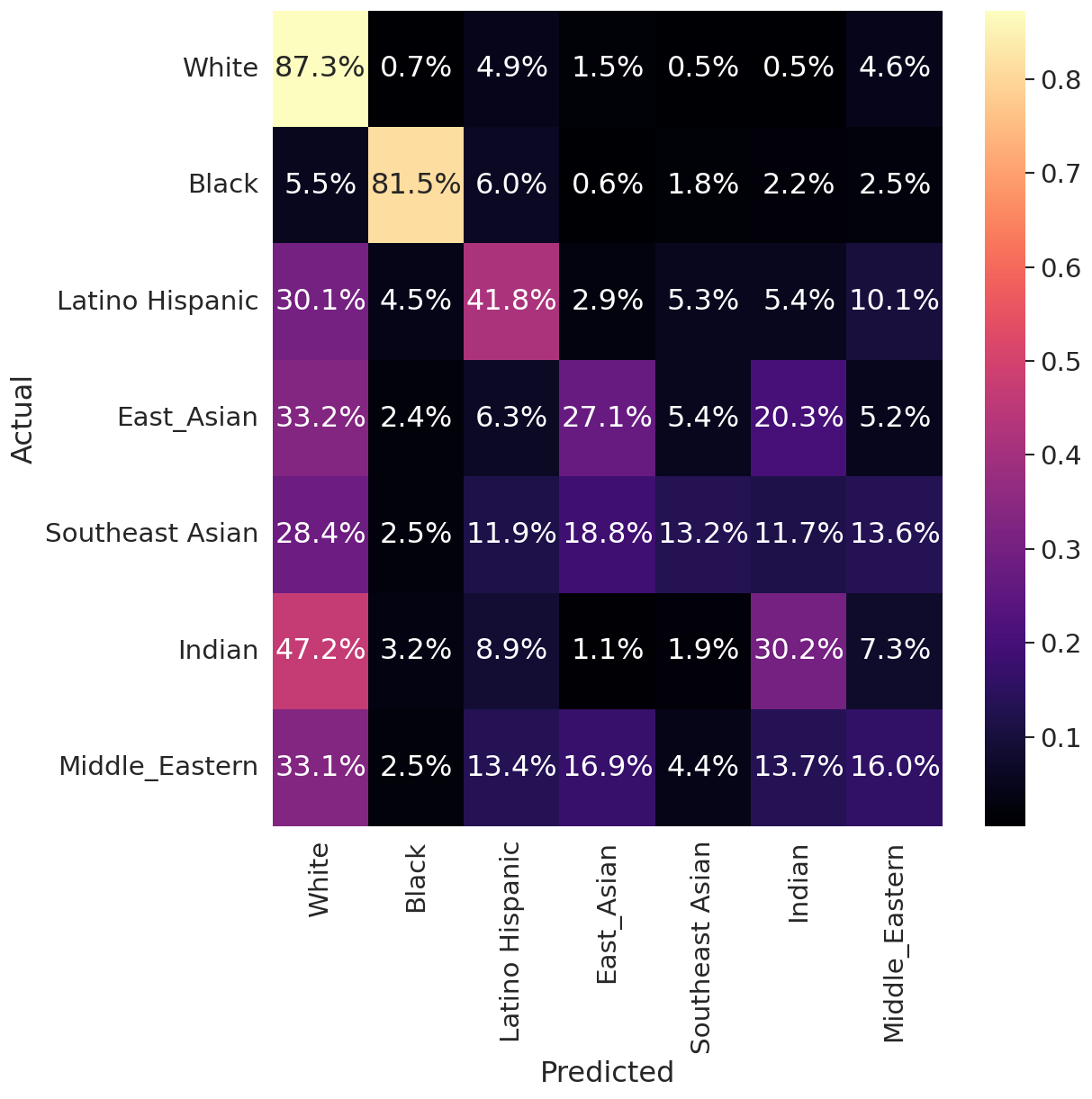}
  \caption{Confusion matrices of the race prediction, obtained through FairFace \cite{fairface}. Left-hand side refers to the original images, right-hand side refers to the images after beautification.}
  \label{fig:ff}
\end{figure}

To deepen the understanding of the results in Tables \ref{tab:deepface} and \ref{tab:ff}, we report the confusion matrices obtained with each of the models in Figures \ref{fig:deepface} and \ref{fig:ff}. The left-hand side of the images depicts the confusion matrix on the original faces whereas the right-hand-side depicts the confusion matrix on the beautified version of the faces. Given the differences in behavior between the two race classification algorithms, we discuss the results separately. With respect to DeepFace (Figure \ref{fig:deepface}), we observe high prediction accuracies on faces labeled as Asian and Black (78.2\% and 80.0\% respective accuracies), and significantly lower on the rest of racial groups (ranging between 39.5\% for Middle Eastern and 69.3\% for White). After beautification, the accuracy on faces labeled as \emph{White} increases (74.6\%), whereas the accuracies on faces with all other labels significantly decrease, mostly due to a significant increase in the misclassification of the images as White (right-most column on the confusion matrix). 

\begin{table}
\caption{Race classification results of the FairFace algorithm, applied to 5,000 images. The first and second columns depict the predicted value of the label \textit{white} in the original \textsc{FairFace} and the beautified \textsc{FairBeauty} datasets, respectively; the third column contains to the race prediction loss, i.e. the difference in the race classification performance between the original and the beautified datasets.}
\label{tab:ff}
\centering
\vspace{6pt}
\def\arraystretch{1.2}
\begin{tabular*}{0.9\textwidth}{lccc}
\toprule
& \multicolumn{1}{c}{Original} \hspace{11pt} & \multicolumn{1}{c}{Beautified} \hspace{11pt} & \multicolumn{1}{c}{True Prediction Loss}\\
\midrule
Race: &  &  & \\
\midrule
\hspace{22pt} Black & $2.64 \pm 0.17$ & $5.98 \pm 0.24$ & -5.99\%\\
\hspace{22pt} East Asian & $30.85 \pm 0.33$ & $32.14 \pm 0.33$ & -1.15\% \\
\hspace{22pt} Indian & $41.98 \pm 0.44$ & $45.00 \pm 0.42$ & -9.25\% \\
\hspace{22pt} Latino & $27.86 \pm 0.26$ & $28.97 \pm 0.41$ & +15.69\%\\
\hspace{22pt} Middle Eastern & $28.58 \pm 0.22$ & $32.08 \pm 0.23$ & -0.5\%\\
\hspace{22pt} Southeast Asian \hspace{11pt} & $27.02 \pm 0.20$ & $27.66 \pm 0.20$ & -1.91\%\\
\hspace{22pt} White & $76.37 \pm 0.42$ & $81.44 \pm 0.36$ & +5.07\%\\
\bottomrule
\end{tabular*}
\end{table}

Regarding the FairFace race classification algorithm (Figure \ref{fig:ff}), we observe in that the highest classification accuracies are obtained for faces labeled as White and Black, with poor performances in the rest of groups (ranging between 15.7\% for Southeast Asian and 40.0\% for Indian). After beautification, the classification performance decreases in all cases, except for faces labeled as White and Latino-Hispanic, which are better classified. Moreover, the probability of misclassifying faces from all racial groups as White significantly increases (first column in the confusion matrix). In general, the FairFace classifier seems to be biased towards the \textit{White} label, misclassifying faces from all other groups as being White even in the original, non-beautified case. This effect is exacerbated after beautification: the accuracy on the \textit{White} label increases (87.3\%), to the detriment of all the other labels except for the Latino-Hispanic label.

Our experiments yield results that, while preliminary, highlight societal and cultural issues that would need deeper investigation. In particular, the \textit{Beautyverse} not only homogenizes the visual aesthetics of faces as reported in \cite{riccio2022}, but seems to make them conform with a canon of beauty of white people. As social media platforms (and the Metaverse) aim to reach a globalized community of users, it is unacceptable that the technologies that populate these platforms replicate  intrinsic and subtle biases that perpetuate historic discrimination and privileges.

\section{Future Work and Conclusion}
As an imminent direction of future work, we plan to investigate the impact of beautification filters on higher resolution images. The images in the  \textsc{FairFace} dataset only have a resolution of approximately 300 pixels, which severely limits the beautification process. We expect the effect of the beauty filters to be significantly more prominent on higher resolution images with more detailed facial features. Hence, while the results of our experiments could be seen as a worst-case scenario, we believe that it would be important to perform a similar study on higher resolution images, which are also expected in the Metaverse.

Quantitatively evaluating the existence of a racial bias in the \textit{Beautyverse} is certainly an interesting and important endeavor. However, to counteract this issue, we need to shed light on the specific features implemented in these filters that contribute to the \emph{whitening} of the faces. We plan to leverage state-of-the-art explainability frameworks (e.g. \cite{gradcam}) to automatically identify the areas in the images that are responsible for the shift in the classification. Explaining these results could lead to two relevant insights: (1) first, it would shed light on the specific features --associated with canons of beauty of white people-- that are embedded in the beautification filters and that are considered desirable on social media; and (2) second, it would enable us to assess to what extent this racial bias is related to the beautification filters and to what extent it is intrinsic to the algorithms that classify face images according to race, as we have observed in the FairFace classifier. 

Moreover, we are not aware of any extensive user study to investigate the existence of racial biases in AR-based beauty filters. If we aim to develop an inclusive Metaverse where anyone is welcome, we believe that these issues need to be fully understood and addressed. While our work has been performed on AR-based filters applied to selfies on social media platforms, we believe that the development of the Metaverse will benefit from an understanding of current uses of AR technologies for self-representation on social media. Thus, the results of our work should be valuable to inform the development of Computer Vision-based technologies for self-representation in the Metaverse. The perpetuation of discriminating and racist ideals of beauty applied to avatars in the Metaverse could lead to significant cultural damage and mental health issues (e.g. dysmorphia, anxiety, depression) that need to be studied, understood and mitigated. The research described in this paper contributes to such an understanding. 

\subsubsection{Acknowledgements} P.R. and N.O. are supported by a nominal grant received at the ELLIS Unit Alicante Foundation from the Regional Government of Valencia in Spain (Convenio Singular signed with Generalitat Valenciana, Conselleria d’Innovació, Universitats, Ciència i Societat Digital, Dirección General para el Avance de la Sociedad Digital). P.R. is also supported by a grant by the Banc Sabadell Foundation. 

\clearpage
\bibliographystyle{splncs04}
\bibliography{egbib}

\begin{thebibliography}{10}
\providecommand{\url}[1]{\texttt{#1}}
\providecommand{\urlprefix}{URL }
\providecommand{\doi}[1]{https://doi.org/#1}

\bibitem{abijaoude2020}
Abi-Jaoude, E., Naylor, K.T., Pignatiello, A.: Smartphones, social media use
  and youth mental health. Canadian Medical Association Journal
  \textbf{192}(6) (2020), \url{https://doi.org/10.1503/cmaj.190434}

\bibitem{metaverse2040}
Anderson, J., Rainie, L.: The metaverse in 2040. Pew Research Center  (2022)

\bibitem{bessiere07}
Bessi{\`e}re, K., Seay, A.F., Kiesler, S.: The ideal elf: Identity exploration
  in world of warcraft. Cyberpsychology \& behavior  \textbf{10}(4),  530--535
  (2007)

\bibitem{ducheneaut09}
Ducheneaut, N., Wen, M.H., Yee, N., Wadley, G.: Body and mind: a study of
  avatar personalization in three virtual worlds. In: Proceedings of the SIGCHI
  conference on human factors in computing systems. pp. 1151--1160 (2009)

\bibitem{higgins87}
Higgins, E.T.: Self-discrepancy: a theory relating self and affect.
  Psychological review  \textbf{94}(3), ~319 (1987)

\bibitem{complex}
Jagota, V.: Why do all the snapchat filters try to make you look white? (Jun
  2016)

\bibitem{fairface}
Karkkainen, K., Joo, J.: Fairface: Face attribute dataset for balanced race,
  gender, and age for bias measurement and mitigation. In: Proceedings of the
  IEEE/CVF Winter Conference on Applications of Computer Vision (WACV). pp.
  1548--1558 (January 2021)

\bibitem{khunger2021}
Khunger, N., Pant, H.: {Cosmetic Procedures in Adolescents: What's Safe and
  What Can Wait}. Indian Journal of Paediatric Dermatology  \textbf{22}(1),
  12--20 (2021). \doi{10.4103/ijpd.IJPD\_53\_20}

\bibitem{552}
Kolesnichenko, A., McVeigh-Schultz, J., Isbister, K.: Understanding emerging
  design practices for avatar systems in the commercial social vr ecology. In:
  Proceedings of the 2019 on Designing Interactive Systems Conference. pp.
  241--252 (2019)

\bibitem{lee21}
Lee, L.H., Braud, T., Zhou, P., Wang, L., Xu, D., Lin, Z., Kumar, A., Bermejo,
  C., Hui, P.: All one needs to know about metaverse: A complete survey on
  technological singularity, virtual ecosystem, and research agenda. arXiv
  preprint arXiv:2110.05352  (2021)

\bibitem{nylon}
Li, S.: The problems with instagram’s most popular beauty filters, from
  augmentation to eurocentrism (Jul 2020)

\bibitem{divine}
Maloney, D.: Mitigating negative effects of immersive virtual avatars on racial
  bias. In: Proceedings of the 2018 Annual Symposium on Computer-Human
  Interaction in Play Companion Extended Abstracts. p. 39–43. CHI PLAY '18
  Extended Abstracts, Association for Computing Machinery, New York, NY, USA
  (2018). \doi{10.1145/3270316.3270599},
  \url{https://doi.org/10.1145/3270316.3270599}

\bibitem{manago08}
Manago, A.M., Graham, M.B., Greenfield, P.M., Salimkhan, G.: Self-presentation
  and gender on myspace. Journal of Applied Developmental Psychology
  \textbf{29}(6),  446--458 (2008)

\bibitem{messinger08}
Messinger, P.R., Ge, X., Stroulia, E., Lyons, K., Smirnov, K., Bone, M.: On the
  relationship between my avatar and myself. Journal For Virtual Worlds
  Research  \textbf{1}(2) (2008)

\bibitem{huffpost}
Mulaudzi, S.: Let's be honest: Snapchat filters are a little racist (Jan 2017),
  \url{https://www.huffingtonpost.co.uk/2017/01/25/snapchat-filters-are-harming-black-womens-self-image\_a\_21658358/}

\bibitem{mummendey95}
Mummendey, H.D.: Psychologie der Selbstdarstellung. Göttingen: Hogrefe (1990)

\bibitem{neely17}
Neely, E.L.: No player is ideal: why video game designers cannot ethically
  ignore players' real-world identities. ACM SIGCAS Computers and Society
  \textbf{47}(3),  98--111 (2017)

\bibitem{vgg-face}
Parkhi, O.M., Vedaldi, A., Zisserman, A.: Deep face recognition. British
  Machine Vision Association  (2015)

\bibitem{riccio2022}
Riccio, P., Psomas, B., Galati, F., Escolano, F., Hofmann, T., Oliver, N.:
  Openfilter: A framework to democratize research access to social media ar
  filters. arXiv preprint arXiv:2207.12319  (2022)

\bibitem{techreview}
Ryan-Mosley, T.: Beauty filters are changing the way young girls see themselves
  (Apr 2021),
  \url{https://www.technologyreview.com/2021/04/02/1021635/beauty-filters-young-girls-augmented-reality-social-media/}

\bibitem{techreview2}
Ryan-Mosley, T.: How digital beauty filters perpetuate colorism (Aug 2021)

\bibitem{gradcam}
Selvaraju, R.R., Das, A., Vedantam, R., Cogswell, M., Parikh, D., Batra, D.:
  Grad-cam: Why did you say that? arXiv preprint arXiv:1611.07450  (2016)

\bibitem{deepface}
Serengil, S.I., Ozpinar, A.: Lightface: A hybrid deep face recognition
  framework. In: 2020 Innovations in Intelligent Systems and Applications
  Conference (ASYU). pp. 23--27. IEEE (2020).
  \doi{10.1109/ASYU50717.2020.9259802},
  \url{https://doi.org/10.1109/ASYU50717.2020.9259802}

\bibitem{2021filtering}
Shein, E.: Filtering for beauty. Communications of the ACM  \textbf{64}(11),
  17--19 (2021)

\bibitem{woodruff18}
Woodruff, A., Fox, S.E., Rousso-Schindler, S., Warshaw, J.: A qualitative
  exploration of perceptions of algorithmic fairness. In: Proceedings of the
  2018 CHI Conference on Human Factors in Computing Systems. p. 1–14. CHI
  '18, Association for Computing Machinery, New York, NY, USA (2018).
  \doi{10.1145/3173574.3174230}, \url{https://doi.org/10.1145/3173574.3174230}

\end{thebibliography}
\end{document}